\documentclass[letterpaper]{article} 
\usepackage{aaai24}  
\usepackage{times}  
\usepackage{helvet}  
\usepackage{courier}  
\usepackage[hyphens]{url}  
\usepackage{graphicx} 
\usepackage{natbib}  
\usepackage{caption} 
\usepackage{algorithm}
\usepackage{algorithmic}

\usepackage{booktabs}
\usepackage{multirow}
\usepackage{pifont}
\usepackage{amsfonts}
\usepackage{amsmath}
\usepackage[algo2e,ruled,noend]{algorithm2e}
\usepackage{color}
\usepackage{colortbl}
\usepackage{bm}

\newcommand{\method}[0]{FLAP }
\newcommand{\methodlong}[0]{\method (\textbf{FL}uctuation-based \textbf{A}daptive Structured \textbf{P}runing)}

\newcommand{\tbf}{\textbf}%
\newcommand{\mbf}{\mathbf}%
\definecolor{gray}{gray}{0.95}

\usepackage{newfloat}
\usepackage{listings}
\DeclareCaptionStyle{ruled}{labelfont=normalfont,labelsep=colon,strut=off} 
\floatstyle{ruled}
\newfloat{listing}{tb}{lst}{}
\floatname{listing}{Listing}
\title{Fluctuation-based Adaptive Structured Pruning for Large Language Models}
\author {
    Yongqi An\textsuperscript{\rm 1, \rm 2},
    Xu Zhao\textsuperscript{\rm 1, \rm 4, }\thanks{Corresponding Author},
    Tao Yu\textsuperscript{\rm 1, \rm 2},
    Ming Tang\textsuperscript{\rm 1, \rm 2},
    Jinqiao Wang\textsuperscript{\rm 1, \rm 2, \rm 3, \rm 4}
}
\affiliations {
    \textsuperscript{\rm 1}Foundation Model Research Center, Institute of Automation, Chinese Academy of Sciences, Beijing, China\\
    \textsuperscript{\rm 2}School of artificial intelligence, University of Chinese Academy of Sciences, Beijing, China\\
    \textsuperscript{\rm 3}Wuhan AI Research, Wuhan, China\\
    \textsuperscript{\rm 4}Objecteye Inc., Beijing, China \\
    \{yongqi.an, xu.zhao, tangm, jqwang\}@nlpr.ia.ac.cn \\ yutao2022@ia.ac.cn
}

\usepackage{bibentry}

\begin{document}

\maketitle

\begin{abstract}

Network Pruning is a promising way to address the huge computing resource demands of the deployment and inference of Large Language Models (LLMs). Retraining-free is important for LLMs' pruning methods. However, almost all of the existing retraining-free pruning approaches for LLMs focus on unstructured pruning, which requires specific hardware support for acceleration. In this paper, we propose a novel retraining-free structured pruning framework for LLMs, named \methodlong. It is hardware-friendly by effectively reducing storage and enhancing inference speed. For effective structured pruning of LLMs, we highlight three critical elements that demand the utmost attention: formulating structured importance metrics, adaptively searching the global compressed model, and implementing compensation mechanisms to mitigate performance loss. First, \method determines whether the output feature map is easily recoverable when a column of weight is removed, based on the fluctuation pruning metric. Then it standardizes the importance scores to adaptively determine the global compressed model structure. At last, \method adds additional bias terms to recover the output feature maps using the baseline values. We thoroughly evaluate our approach on a variety of language benchmarks. Without any retraining, our method significantly outperforms the state-of-the-art methods, including LLM-Pruner and the extension of Wanda in structured pruning. The code is released at \textit{\url{https://github.com/CASIA-IVA-Lab/FLAP}}.

\end{abstract}

\section{Introduction}\label{sec:intr}

Large Language Models (LLMs)~\cite{brown2020gpt3,touvron2023llama,zhang2022opt,scao2022bloom} have recently achieved outstanding performance across various language benchmarks in NLP~\cite{gpt_bar,bubeck2023sparks,wei2022emergent}, spurring a large number of open-source applications~\cite{taori2023alpaca,anand2023gpt4all,richards2023auto}. These remarkable capabilities typically come with a huge-scale model size with high inference costs. This makes it harder for more people to benefit from LLMs. Due to the computational resource constraints, most of the model compression methods in the pre-LLM era are no longer feasible for LLMs. Model compression methods for LLMs to date focus on model quantization~\cite{dettmers2022llmint8,xiao2022smoothquant,frantar2023gptq,dettmers2023spqr} and unstructured pruning~\cite{sun2023simple, frantar2023sparsegpt}.

Structured pruning~\cite{he2023structured}, which prunes entire rows or columns of weights, offers a promising solution to the deployment challenges of LLMs. Unlike unstructured pruning, structured pruning reduces both parameters and inference time without relying on specific hardware, making it more widely applicable~\cite{anwar2017structured}. For effective structured pruning, it's crucial to have a metric that captures the collective significance of an entire row or column. However, current unstructured pruning techniques for LLMs, as seen in methods like \cite{sun2023simple, frantar2023sparsegpt}, primarily focus on the importance of individual elements of each row in isolation. This absence of structured metrics that evaluate entire rows or columns makes them less suitable for structured pruning. The recent LLM-Pruner~\cite{ma2023llmv3} attempted structured pruning for LLMs, but its dependence on LoRA fine-tuning~\cite{hu2021lora} creates a tough trade-off between high computation and effective pruning, limiting its use in larger models.

Pruning essentially involves two key aspects: discovering redundancy and recovering performance. For an effective structured pruning method tailored to LLMs, three fundamental criteria must be satisfied: a) a structured importance metric to discover structured redundancy; b) a mechanism for adaptively searching the optimal global compression model structure; and c) a compensation strategy to minimize performance degradation.

In response to these three essential criteria, we introduce \methodlong, a novel structured pruning framework. We find that certain channels of hidden state features exhibit structured sample stability. This observation enables us to compensate for bias within the model using baseline values. Specifically, we design a structured pruning metric that estimates the fluctuation of each input feature relative to the baseline value, utilizing a set of calibration samples. This metric assists in determining whether the output feature map can be recovered when a column of the weight matrix is removed. We then standardize these fluctuation metric scores across layers and modules separately, allowing for the adaptive determination of the global compressed model structure. Finally, \method employs the baseline values to add additional biases, recovering the output feature maps for the corresponding layers. Remarkably, our method avoids the need for the retraining process and requires only a single forward pass for both pruning and bias compensation, thereby maintaining low memory overhead.

We evaluate the effectiveness of \method on the LLaMA model family, and \method achieves remarkable performance on a variety of language benchmarks. Impressively, without any retraining, our method significantly outperforms the state-of-the-art methods, including LLM-Pruner and the extension of Wanda in structured pruning.

Our main contributions are listed as follows:
\begin{itemize}
    \setlength{\itemsep}{0pt}
    \setlength{\topsep}{0pt}
    \setlength{\parsep}{0pt}
    \item We propose a novel retraining-free structured pruning framework for LLMs named FLAP. To our best knowledge, this is the first work that identifies the characteristic of structured sample stability in LLMs.
    \item The proposed framework uses a bias compensation mechanism, a pruning performance recovery method that does not require retraining. This mechanism yields greater benefits, especially under large pruning ratios. 
    \item Our method achieves remarkable performance on a variety of language benchmarks and outperforms the state-of-the-art method without any retraining.
\end{itemize}

\section{Related Works}\label{sec:related}

\subsection{Network Pruning Methods}\label{sec:related:pruning}

Network pruning is a model compression technique that identifies and eliminates redundancy in the structure or parameters of a neural network, based on specific pruning metrics, and incorporates methods to recover model performance~\cite{lecun1990obd,hassibi1993obs,han2015magnitude}. Pruning methods fall into two categories: unstructured pruning and structured pruning. Unstructured pruning is performed at the individual weight level, allowing for a large sparsity but failing to achieve real inference acceleration or storage reduction~\cite{zafrir2021prune,han2016deepcompress}. Within unstructured pruning, there exists a specialized variant known as semi-structured pruning. This approach enforces exactly N non-zero values in each block of M consecutive weights~\cite{zhou2021learning}. This approach has gained traction recently, particularly with support on newer NVIDIA hardware~\cite{mishra2021accelerating}. Structured pruning, by contrast, operates on entire rows or columns of weights, providing a more hardware-friendly solution that reduces storage requirements and enhances inference speed~\cite{xia2022structured,molchanov2016prune}. 

However, conventional structured pruning methods typically rely on retraining (sometimes iteratively) to regain the performance of the pruned model~\cite{han2015magnitude, tan2020dropnet,han2016deepcompress}. Such methods pose scalability challenges for billion-scale LLMs due to constraints on memory and computational resources. Therefore a retraining-free structured pruning method for LLMs is very critical.

\subsection{Large Language Model Compression}\label{sec:related:llm_compress}

Large Language Models usually consist of billions of parameters, and their gradient backpropagation and training stage require large amounts of memory and computational resources. Consequently, many conventional model compression techniques have become infeasible for LLMs~\cite{frantar2023sparsegpt}. For instance, knowledge distillation~\cite{hinton2015distilling}, once a practical approach, now faces implementation challenges due to high training costs. Existing compression methods for LLMs mainly include post-training quantization~\cite{dettmers2022llmint8,xiao2022smoothquant,frantar2023gptq,dettmers2023spqr} and post-training pruning~\cite{sun2023simple, frantar2023sparsegpt}. Our method also falls into the category of post-training pruning. It utilizes bias compensation to recover model performance, effectively avoiding the high computational cost of retraining. Unlike the past post-training pruning methods, our method is designed for the features of structured pruning of LLMs.

\subsection{Properties of LLMs}\label{sec:related:properties}

Our work is related to the distinct properties of Large Language Models (LLMs) that have inspired various model compression techniques~\cite{sun2023simple,dettmers2023spqr,dettmers2022llmint8}. Dettmers et al.~\cite{dettmers2022llmint8} observed the emergence of channels with abnormally large magnitudes in the hidden state features of LLMs once they exceed a certain parameter scale (e.g., 6B). They suggest that this is the reason why existing quantization methods fail on LLMs. In response, they introduced a novel mixed-precision quantization technique. Contrary to the focus of previous work on the outlier magnitudes in LLMs, our research pivots towards investigating the structured stability within the channels of input features in these models. In our study, we find that certain channels within the hidden state features demonstrate consistent structured sample stability. This discovery offers invaluable insights for crafting structured post-training pruning algorithms, laying the foundation for the method we present in this paper.

\begin{figure*}[t]
    \centering
    \includegraphics[width=0.86\linewidth]{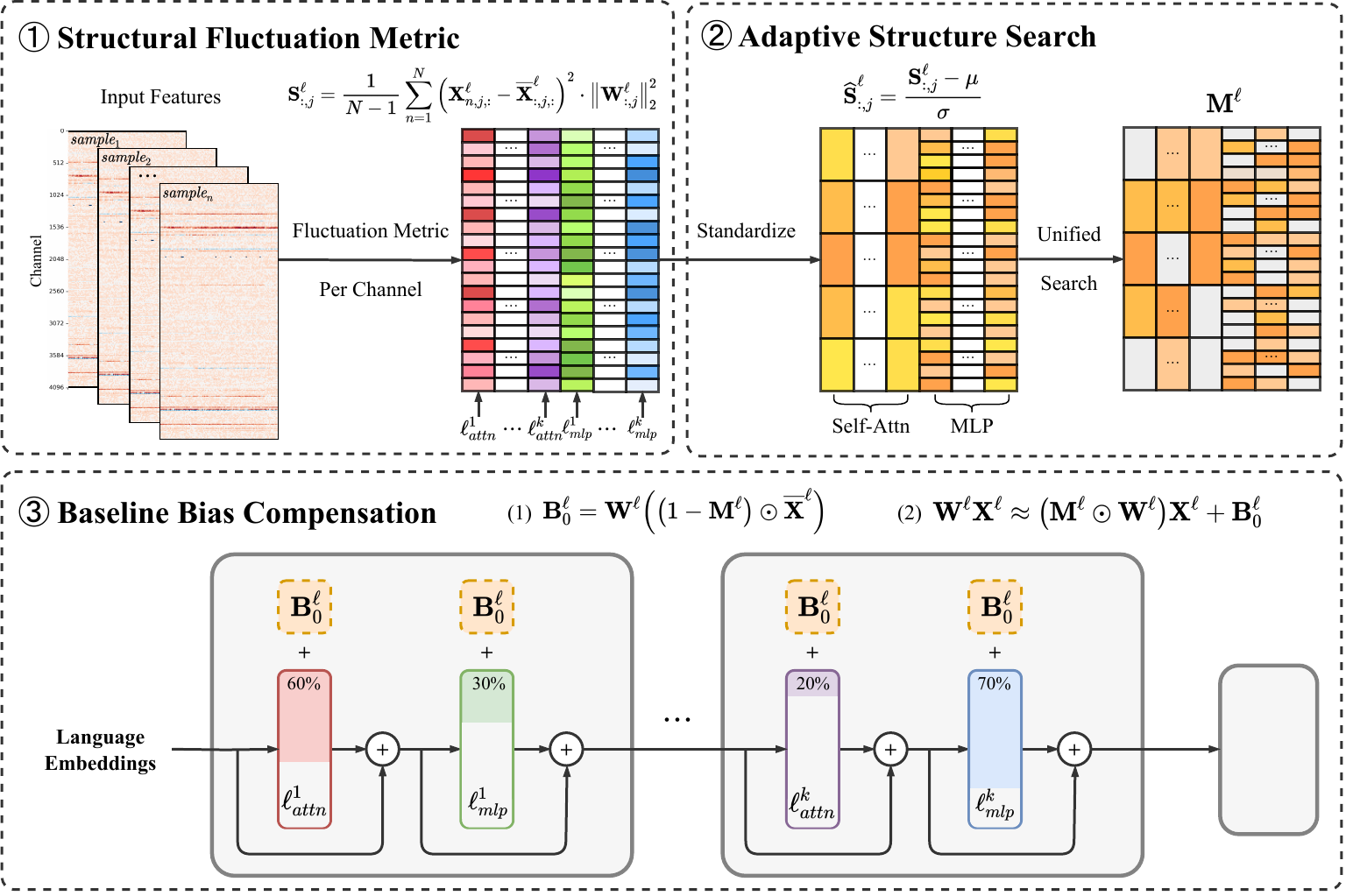}
    \caption{Framework of the proposed FLAP. \ding{172}Measure the fluctuation of each channel across different layers and modules using calibration data; \ding{173}Standardize these fluctuation measures for a unified search method; \ding{174}Implement adaptive pruning ratios for each layer and module, employing bias compensation to restore model performance.}
    \label{fig:overview}
     \vspace{-0.5cm}
\end{figure*}

\section{Preliminaries}\label{sec:pre}

\subsection{Layer-Wise Pruning} 

Given the computational constraints, globally solving the pruning problem for Large Language Models (LLMs) is challenging. Layer-wise pruning becomes a practical solution under these constraints. Following this notion, SparseGPT~\cite{frantar2023sparsegpt} demonstrated that the challenge of unstructured pruning for LLMs can be tackled by decomposing it into individual layer-wise subproblems. This principle can be seamlessly extended to structured pruning within LLMs. The quality of solutions to these layer-wise subproblems can be evaluated based on the $\ell_2$-error. Given an input $\mathbf{X}^{\ell}$ of shape $\left(N, C_{in}, L\right)$ where $N$ and $L$ represent batch and sequence dimensions respectively, and a weight $\mathbf{W}^{\ell}$ of shape $\left(C_{out}, C_{in}\right)$, the $\ell_2$-error for structured pruning can be defined as:
\begin{equation}
    \text{argmin}_{\mathbf{M}^{\ell} \in \mathbb{R}^{C_{in}}, \widehat{\mathbf{W}}^{\ell}}||\mathbf{W}^{\ell} \mathbf{X}^{\ell}-(\mathbf{M}^{\ell} \odot \widehat{\mathbf{W}}^{\ell}) \mathbf{X}^{\ell}||_2^2
    \label{eq:layer-wise}
\end{equation}

\noindent{where $\mathbf{M}^{\ell} \in \mathbb{R}^{C_{in}}$ represents the mask vector corresponding to the input channels, this vector mirrors whether each input channel is pruned or not. For the self-attention modules, these input channels are pruned in groups typically with sizes like group$\_$size=128. The term $\widehat{\mathbf{W}}^{\ell}$ denotes the possibly updated weights for the pruned layer. The notation $|| \cdot ||_2^2$ represents the $\ell_2$-error.}

\subsection{Local Pruning Metric Challenges}

Regarding Eq.~\eqref{eq:layer-wise}, the existing methods can be broadly categorized into two primary approaches: low-damage and easy-recoverability. These correspond to the core principles of OBD~\cite{lecun1990obd} and OBS~\cite{hassibi1993obs}, respectively. To illustrate, Wanda~\cite{sun2023simple} uses a localized low-damage pruning metric to minimize harm to each layer's output features. In contrast, SparseGPT~\cite{frantar2023sparsegpt} employs an easy-recoverability metric, aiming to identify components that other weights can compensate for during pruning. These approaches are insightful but tend to focus on the importance of individual elements in the weight matrix, neglecting the broader structured context. Such an atomistic approach is misaligned with structured pruning's requirements, which demand a more global perspective that captures the collective importance of entire rows or columns in the matrix.

\section{Methodology}\label{sec:method}

In this section, we introduce FLAP, our proposed approach to structured pruning for Large Language Models (LLMs). \method encompasses three key components: Baseline Bias Compensation, Structured Fluctuation Metric, and Adaptive Structure Search. The overview of our method is presented in Figure~\ref{fig:overview}.

\subsection{Baseline Bias Compensation}\label{sec:method:bias}

In the context of structured pruning, the output of the layers of the uncompressed model can be decomposed into:
\begin{equation}
    \mathbf{W}^{\ell} \mathbf{X}^{\ell} = \underbrace{(\mathbf{M}^{\ell} \odot \mathbf{W}^{\ell}) \mathbf{X}^{\ell}}_{\text{retained}} + \underbrace{((1 - \mathbf{M}^{\ell}) \odot \mathbf{W}^{\ell}) \mathbf{X}^{\ell}}_{\text{removed}}
\end{equation}

The objective of structured pruning is to minimize the impact introduced by $\Delta Y^{\ell} = ((1 - \mathbf{M}^{\ell}) \odot \mathbf{W}^{\ell}) \mathbf{X}^{\ell}$ in the overall output feature map, thereby reducing the reconstruction error for each layer. For structured pruning of LLMs, the constraints are stronger, so the latter components cannot be simply removed. Therefore, a compensatory mechanism is essential to recover the model's performance while adhering to the pruning structure.

We add an additional bias term to compensate for the damage inflicted on the output feature maps by the removed components. This bias term is designed to mitigate the reconstruction error introduced by the pruning process, allowing the pruned model to maintain high performance. In particular, we construct the bias term based on the baseline value, $\overline{\mathbf{X}}^{\ell}_{:,j,:}$, which represents the average of the $j$-th channel for all samples in layer $l$. As detailed in the following section, our empirical findings validate the effectiveness and feasibility of this compensatory approach. Specifically, the formulation for the baseline value is as follows:
\begin{equation}
    \overline{\mathbf{X}}^{\ell}_{:,j,:} = \frac{1}{NL} \sum_{n=1}^{N} \sum_{k=1}^{L} \mathbf{X}^{\ell}_{n,j,k}
\end{equation}

Once the mask $\mathbf{M}^{\ell}$ is established, the baseline value for the pruned channel can be seamlessly translated into the bias term for the linear layer as follows:
\begin{equation}
\begin{aligned}
    \mathbf{B}^{\ell}_0 & = \mathbf{W}^{\ell} ((1 - \mathbf{M}^{\ell}) \odot \overline{\mathbf{X}}^{\ell}) \\
    \mathbf{W}^{\ell} \mathbf{X}^{\ell} & \approx (\mathbf{M}^{\ell} \odot \mathbf{W}^{\ell}) \mathbf{X}^{\ell} + \mathbf{B}^{\ell}_0
    \label{eq:bbc}
\end{aligned}
\end{equation}

\noindent{where $\mathbf{B}_0$ represents the bias of linear layer, which has a shape of $(C_{out}, )$, and $\overline{\mathbf{X}}^{\ell}$ is a one-dimensional vector with dimensions $(C_{in},)$.}

\subsection{Structured Fluctuation Metric}\label{sec:method:metric}

\begin{figure}[ht!]
    \centering 
  \includegraphics[width=0.95\linewidth]{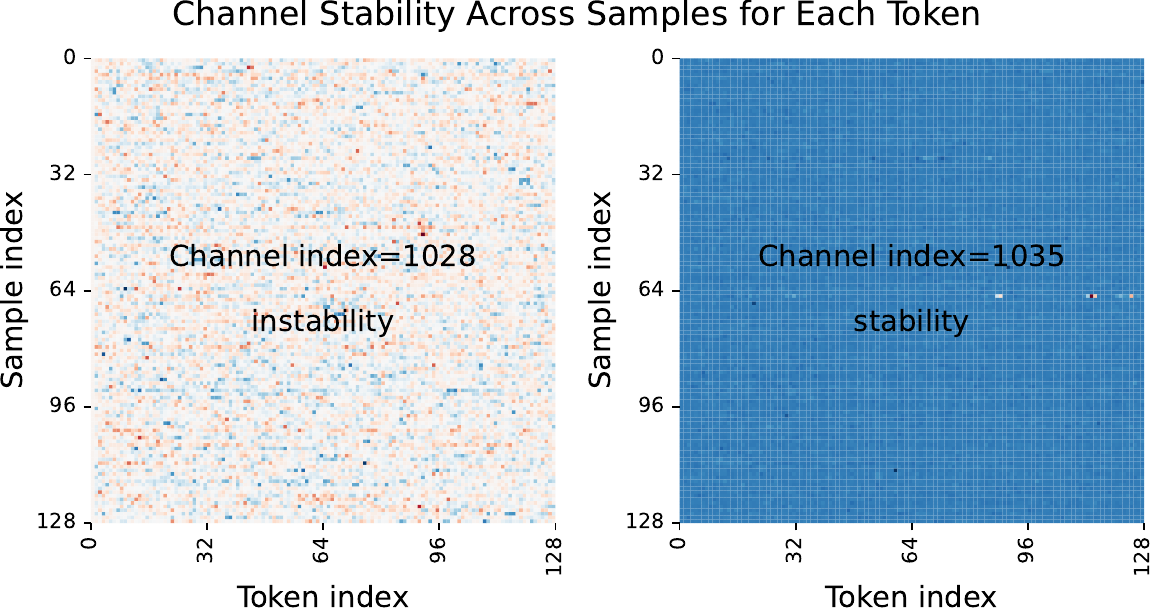}
\caption{Certain channels of hidden state features exhibit structured sample stability. The left shows a channel with noticeable variations across samples, indicating low stability. The right displays a stable pattern common in many LLaMa channels.}
\label{fig:features_stable}
\end{figure}

Motivated by the observations from Figure~\ref{fig:features_stable}, we note that certain channels of the hidden state features exhibit a low variation across different samples. This low fluctuation indicates that if their corresponding input feature channels are pruned, the resulted change in the output feature map can be effectively counterbalanced by the baseline value.

As illustrated in Eq.~\eqref{eq:bbc}, the structured easy-recoverability metric seeks to evaluate the impact on the output feature map when an input channel is substituted with its baseline value. A straightforward approach would involve individually substituting each input channel with its baseline value for the calibration samples and then computing the $\ell_2$-error between the output feature maps before and after this replacement.

However, such a method poses a significant computational challenge and is impractical for LLMs. To address this, we introduce an approximate metric for structured recoverability, which termed the "fluctuation metric". Specifically, we compute the sample variance of each input feature and weight it with the squared norm of the corresponding column of the weight matrix. Concretely, the score for the group of weight $\mathbf{W}^{\ell}_{:,j}$ is defined by:
\begin{equation}
\mathbf{S}^{\ell}_{:,j} = \frac{1}{N-1} \sum_{n=1}^N (\mathbf{X}^{\ell}_{n,j,:} - \overline{\mathbf{X}}^{\ell}_{:,j,:})^2  \cdot ||\mathbf{W}^{\ell}_{:,j}||^2_2
\label{eq:fluc}
\end{equation}

\noindent{where $||\mathbf{W}^{\ell}_{:,j}||^2_2$ denotes the squared norm of $j$-th column of the weight matrix. $\frac{1}{N-1} \sum_{n=1}^N (\mathbf{X}^{\ell}_{n,j,:} - \overline{\mathbf{X}}^{\ell}_{:,j,:})^2$ represents the sample variance of the $j$-th channel of the input feature of layer $\ell$ under $N$ calibration samples. The denominator here is $\frac{1}{N-1}$. This correction is known as the Bessel correction and is used for unbiased estimation of the overall variance.}

\subsection{Adaptive Structure Search}\label{sec:method:adaptive}

\begin{figure}[h]
    \centering 
  \includegraphics[width=0.95\linewidth]{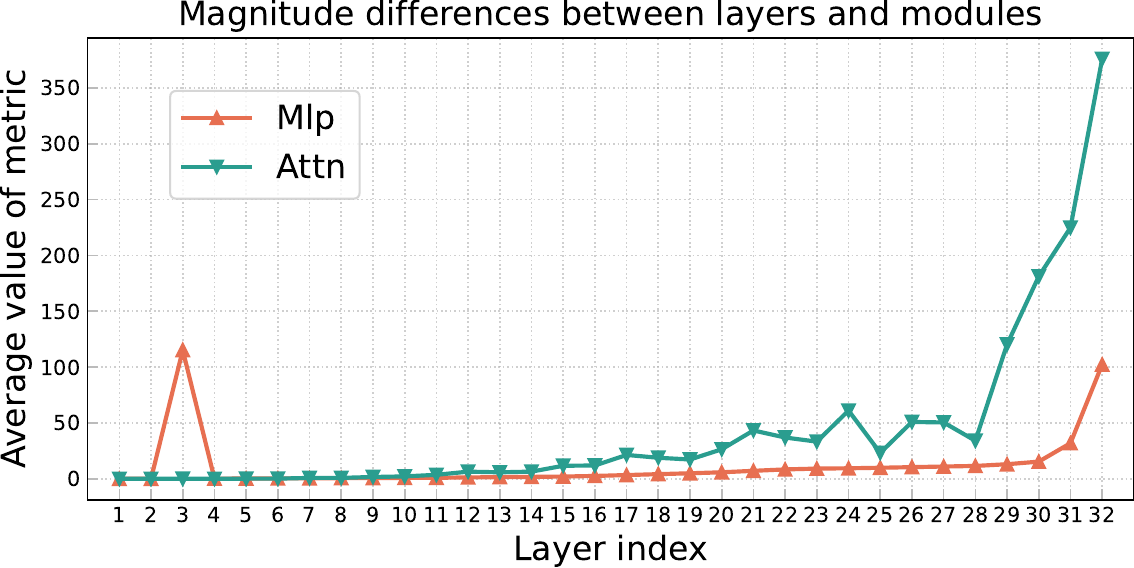}
\caption{Comparison of the average value of the fluctuation metric across different layers for different modules.}
\label{fig:mean_layer_metric}
\end{figure}

The central challenge in layer-wise pruning revolves around adaptively searching the global compression model structures. Unifying different layers and modules without distinction can critically degrade performance. This issue arises because the magnitudes of the metrics across layers and modules vary greatly~\cite{shi2023upop}. Figure~\ref{fig:mean_layer_metric} demonstrates this by showing the mean values of the fluctuation metric for different modules in different layers.

To ensure a consistent comparison of scores across different layers and modules, we standardize the metric distributions for each layer to a common mean and standard deviation. As defined in Eq.~\eqref{eq:fluc}, the fluctuation metric captures the absolute variation in the output feature map when input features are replaced with their baseline values. In contrast, the standardized metric reflects the relative variation in the output feature map resulting from this replacement, making it suitable for a structured unified search. The standardized metric, denoted as, is formulated as follows:
\begin{equation}
\widehat{\mathbf{S}}^{\ell}_{:,j} = (\mathbf{S}^{\ell}_{:,j}-\mathbb{E}[\mathbf{S}^{\ell}_{:,j}]) /(\mathbb{E}[[\mathbf{S}^{\ell}_{:,j}-\mathbb{E}[\mathbf{S}^{\ell}_{:,j}]]^2])^{\frac{1}{2}}
\end{equation}

\noindent{where $\mathbb{E}[\mathbf{S}^{\ell}_{:,j}]$ represents the expected value (or mean) of the vector $\mathbf{S}^{\ell}_{:,j}$. $(\mathbb{E}[[\mathbf{S}^{\ell}_{:,j}-\mathbb{E}[\mathbf{S}^{\ell}_{:,j}]]^2])^{\frac{1}{2}}$ represents the square root of the variance, which is the standard deviation.}

\section{Experiments}\label{sec:exp}

\subsection{Experimental Settings}\label{sec:exp:set}
We conduct experiments on the LLaMA model family (LLaMA-7B/13B/30B/65B) to evaluate the efficacy of \method. Our evaluation focuses on language modeling performance on the WikiText2~\cite{merity2016pointer} validation set and zero-shot performance across seven common sense benchmarks using the EleutherAI LM Harness~\cite{eval-harness}\footnote{https://github.com/EleutherAI/lm-evaluation-harness}. We compare \method against two previous pruning methods: Wanda-sp and LLM-Pruner. We generalize Wanda to structured pruning and name it as Wanda-sp. Detailed experimental settings, model descriptions, and evaluation protocols are provided in the Appendix A.


\begin{table}[ht]
\centering
\renewcommand{\arraystretch}{1.05}
\small
\begin{tabular}{@{}l|c|cccc@{}}
\toprule 
    \multirow{2}{*}{Method} & \multirow{2}{*}{\shortstack{Pruning \\ Ratio}} & \multicolumn{4}{c}{LLaMA} \\
    \cmidrule{3-6}
    & & 7B & 13B & 30B & 65B \\
    \hline
      Dense  & 0$\%$ & 12.62 & 10.81 & 9.11 & 8.21 \\
    \hline
    Wanda-sp & \multirow{4}{*}{20$\%$}  & 22.12 & 16.83 & 11.66 & 11.76\\
    LLM-Pruner &  & 19.77 & 16.01 & - & -\\
    LLM-Pruner* &  & 17.37 & 15.18 & - & -\\
    \method (Ours) &  & \tbf{14.62} & \tbf{13.66} & \tbf{10.86} & \tbf{9.79}\\
    \hline
    Wanda-sp & \multirow{2}{*}{30$\%$} & 38.88 & 22.89 & 14.90 & 14.64\\
    \method (Ours) &  & \tbf{17.62} & \tbf{15.65} & \tbf{12.49} & \tbf{10.90}\\
    \hline
    Wanda-sp & \multirow{4}{*}{50$\%$} & 366.43 & 160.49 & 67.46 & 42.85 \\
    LLM-Pruner &  & 112.44 & - & - & -\\
    LLM-Pruner* &  & 38.12 & - & - & -\\
    \method (Ours) &  & \tbf{31.80} & \tbf{24.20} & \tbf{19.36} & \tbf{15.30}\\
\bottomrule
\end{tabular}
\caption{WikiText2 validation perplexity of pruning methods for LLaMA model family. * means with LoRA fine-tuning.}
\label{tab:ppl_results}
\end{table}

\subsection{Language Modeling}\label{sec:exp:lm}

\subsubsection{Performance Comparisons.}

For each of the LLaMA models, we present results at three distinct pruning ratios, as detailed in Table~\ref{tab:ppl_results}. Notably, \method significantly outperforms the other methods, achieving this superiority without any retraining. As the pruning ratio increases, the performance advantage of \method becomes more significant. To illustrate, consider the LLaMA-7B model: at a 50$\%$ pruning ratio, the LLM-Pruner exhibits a perplexity of 130.97, which improves to 39.02 after LoRA fine-tuning. In stark contrast, \method efficiently identifies sparse networks that yield a perplexity of 31.80, and remarkably, this is achieved without any retraining.

\subsubsection{Remark.}

The \method method, which requires no retraining, consistently outperforms the LLM-Pruner, even when the latter is fine-tuned with LoRA. Eq~\eqref{eq:bbc} offers insight into the potential reason for this superior performance. In \method, the baseline bias $\mathbf{B}_0$ is effectively treated as a low-rank component with a rank of $r=1$. Within the pruning framework of \method, bias compensation plays a pivotal role, serving a function similar to that of LoRA fine-tuning. This compensation helps to effectively recover the model’s performance after pruning.

\begin{table*}[t]
    \centering
    \resizebox{0.9\linewidth}{!}{
    \begin{tabular}{l|c|ccccccc|c}
        \toprule
        Method  & Pruning Ratio & BoolQ & PIQA & HellaSwag & WinoGrande & ARC-e & ARC-c & OBQA & Average \\
        \midrule
        LLaMA-7B  & 0$\%$ & 73.18 & 78.35 & 72.99 & 67.01 & 67.45 & 41.38 & 42.40 & 63.25 \\
        \midrule
        Wanda-sp  & \multirow{4}{*}{20$\%$} & \tbf{71.25} & \tbf{77.09} & \tbf{72.77} & \underline{67.09} & \tbf{71.09} & \tbf{42.58} & \tbf{41.60} & \tbf{63.35} \\
        LLM-Pruner &  & 59.39 & 75.57 & 65.34 & 61.33 & 59.18 & 37.18 & 39.80 & 56.82 \\
        LLM-Pruner (w/ LoRA) &  & 69.54 & 76.44 & 68.11 & 65.11 & 63.43 & 37.88 & \underline{40.00} & 60.07 \\
        \method (Ours)  &  & \underline{69.63} & \underline{76.82} & \underline{71.20} & \tbf{68.35} & \underline{69.91} & \underline{39.25} & 39.40 & \underline{62.08} \\
        \midrule
        Wanda-sp  & \multirow{4}{*}{50$\%$} & 50.58 & 55.01 & 29.56 & 51.78 & 31.27 & 23.04 & 23.60 & 37.83 \\
        LLM-Pruner &  & 52.57 & 60.45 & 35.86 & 49.01 & 32.83 & 25.51 & 34.80 & 41.58 \\
        LLM-Pruner (w/ LoRA) &  & \tbf{61.47} & \tbf{68.82} & \underline{47.56} & \underline{55.09} & \underline{46.46} & \underline{28.24} & \underline{35.20} & \underline{48.98} \\
        \method (Ours) &  & \underline{60.21} & \underline{67.52} & \tbf{52.14} & \tbf{57.54} & \tbf{49.66} & \tbf{29.95} & \tbf{35.60} & \tbf{50.37} \\
        \bottomrule
    \end{tabular}
    }
    \caption{Zero-shot performance of the compressed LLaMA-7B. \tbf{Bold} results highlight the best performance. \underline{Underscored} results denote the second-best performance for each pruning ratio.} 
    \label{tab:zero-shot}
    \vspace{-3mm}
\end{table*}

\subsubsection{Different Pruning Ratio.}

\begin{figure}[ht!]
    \centering %
  \includegraphics[width=0.95\linewidth]{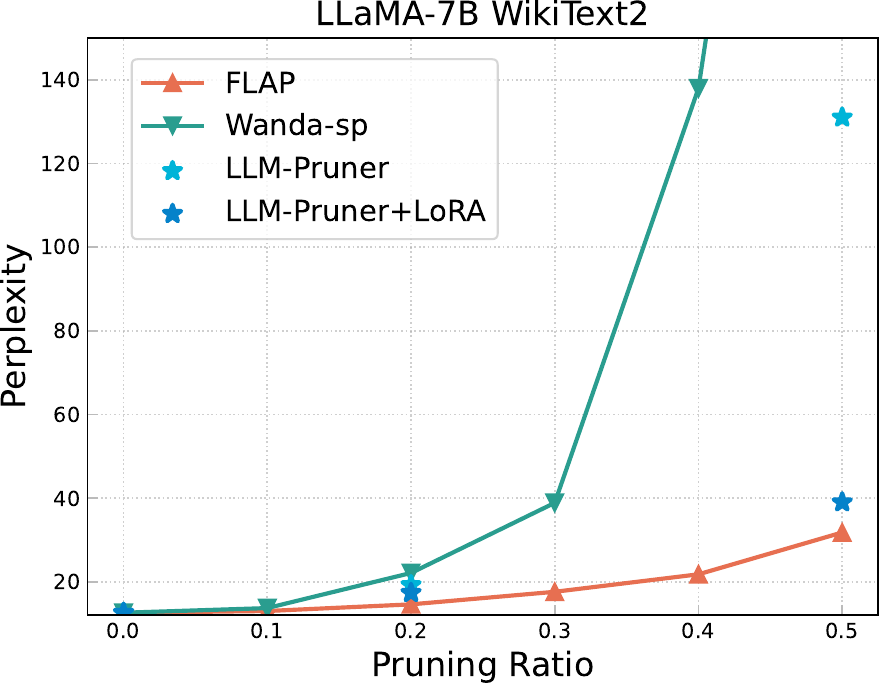}
  \vspace{-0.3cm}
\caption{Results among \method and other structured pruning methods at varying pruning ratios on the LLaMA-7B WikiText2 dataset.}
\label{fig:pruning_ratio}
\end{figure}

We evaluated the performance of each structured pruning method at various pruning ratios. As depicted in Figure~\ref{fig:pruning_ratio}, \method demonstrates remarkable stability in maintaining its performance as the pruning ratio increases. In contrast, Wanda-sp exhibits a sharp decrease in performance as the pruning ratio rises. Meanwhile, LLM-Pruner requires LoRA fine-tuning to maintain acceptable performance when the pruning ratio is increased to levels like 50$\%$.

\subsection{Zero-shot Tasks Performance}\label{sec:exp:zero-shot}

We assessed the zero-shot capability of the pruned model across seven downstream tasks. As illustrated in Table~\ref{tab:zero-shot}, our method consistently outperforms LLM-Pruner with LoRA Fine-Tuning, achieving superior performance across varying pruning ratios, all without the need for retraining. At a 20$\%$ pruning ratio, Wanda-sp exhibits remarkable zero-shot capabilities, even surpassing the performance of the original, unpruned model. This suggests the presence of structured redundancy within LLMs that can be pruned away without necessitating retraining, thereby potentially enhancing model efficiency. However, when the pruning ratio is increased to 50$\%$, the performance of Wanda-sp suffers a significant degradation. In stark contrast, our method continues to excel, maintaining a clear advantage over other approaches. This finding demonstrates the efficacy of our structured pruning method in preserving the generalization capabilities of large language models (LLMs), even under stringent pruning conditions.

\subsection{Ablation Study}\label{sec:exp:ablation}

We systematically examine three fundamental components of the \method method: the pruning metric, the global compression structure, and bias compensation. Additionally, we evaluate the robustness of our pruning approach in relation to calibration samples.

\subsubsection{Pruning Metric.} Both the pruning metric and compressed model structure are critical factors in the pruning process. \method is specifically designed to address these two dimensions in the structured pruning of Large Language Models (LLMs). To evaluate their effectiveness, we conducted experiments employing various structured pruning metrics and global compression structures. 

We investigated three structured pruning metrics in this study: 1) Weighted Input Feature Norm (WIFN), a low-damage metric assessing the effect of weight columns on the output feature map; 2) Input Feature Variance (IFV), used to gauge the variability among input features; and 3) Weighted Input Feature Variance (WIFV), utilized by \method to assist in determining the potential for recovery of the output feature map after a column of the weight matrix is removed.

To underscore the importance of global adaptive compression structure, we defined four configurations: 'UL-UM' (Uniform across Layers and Modules, employed in unstructured pruning for LLMs like Wanda); 'UL-MM' (Uniform across Layers, Manual ratio for Modules); 'AL-MM' (Adaptive across Layers, Manual for Modules); and 'AL-AM' (Adaptive across both Layers and Modules), the structure chosen by \method. Results in this section include bias compensation, with bias-compensated ablation experiments detailed later.

\begin{table*}[ht!]
 \centering
 \renewcommand{\arraystretch}{1.4}
\small
\begin{tabular}{lccccc}
                \hline 
          & \multicolumn{4}{c}{Compressed model structure} \\
    \cmidrule{2-5}
    Pruning Metric &   UL-UM & UL-MM & AL-MM & AL-AM \\ \hline 
    WIFN: $\sum_{i=1}^{C_{out}}  ||\mbf{X}^{\ell}_{j}||_2 \cdot |\mbf{W}^{\ell}_{ij}|$   & 84.79  & 128.75  & \underline{34.50} &  \tbf{34.09} \\ 
    IFV: $\frac{1}{N-1} \sum_{n=1}^N (\mathbf{X}^{\ell}_{n,j,:} - \overline{\mathbf{X}}^{\ell}_{:,j,:})^2$ & \underline{55.41} & 48.87 & 35.72 &  \tbf{33.33} \\
    WIFV: $\frac{1}{N-1} \sum_{n=1}^N (\mathbf{X}^{\ell}_{n,j,:} - \overline{\mathbf{X}}^{\ell}_{:,j,:})^2  \cdot ||\mathbf{W}^{\ell}_{:,j}||^2_2$ & 57.57 & \underline{38.31} & 34.82 &  \underline{\tbf{31.80}}\\
    \hline 
    \end{tabular}
\caption{Ablation on pruning metric and compressed model structure. \tbf{Bold} results denote the best compressed model structure found for each pruning metric. \underline{Underscored} results indicate the best pruning metric found for each compressed model structure.}
\label{tab:ablate_metric_structure}
\end{table*}

In our experiments, we structurally pruned the LLaMA-7B model with a 50$\%$ pruning ratio and evaluated the model using the perplexity metric on the WikiText2 dataset. The detailed results are presented in Table~\ref{tab:ablate_metric_structure}. Notably, the most effective pruning model was obtained using the default configuration of \method, achieving a perplexity of 31.80. The AL-AM global adaptive compression structure consistently outperformed other configurations under all evaluated pruning metrics, thereby effectively validating our proposed Adaptive Structure Search strategy. When analyzing the effectiveness of different global compression structures, we observed that various metrics present distinct strengths and weaknesses. Nevertheless, our proposed WIFV structured pruning metric displayed superior adaptability to the global compression structure.

\subsubsection{Baseline Bias Compensation.} 
\begin{figure}[t]
    \centering %
  \includegraphics[width=0.9\linewidth]{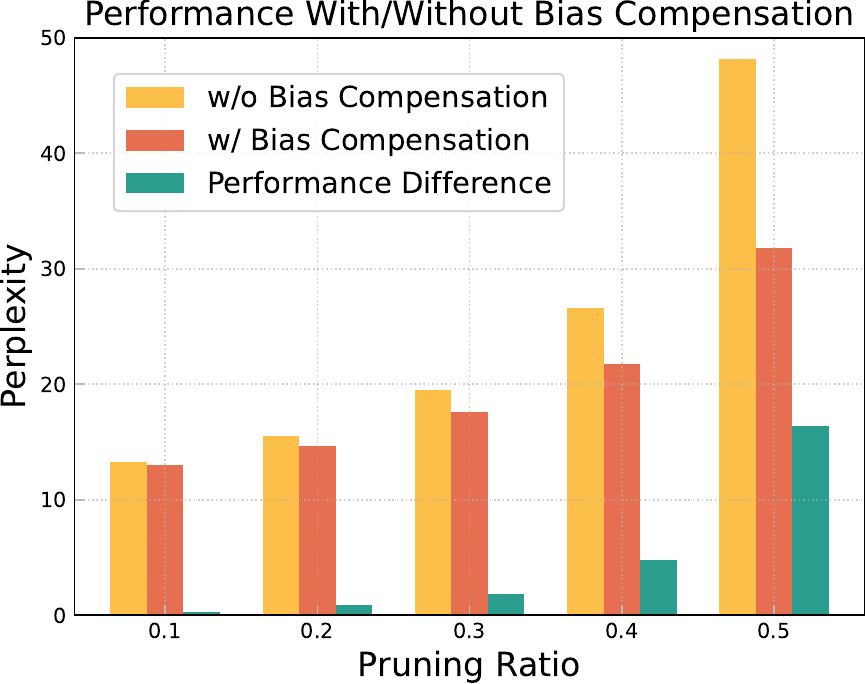}
\vspace{-0.23cm}
\caption{Performance comparison of the model with and without Bias Compensation at various pruning ratios. The yellow and orange bars represent the Perplexity of the model without and with Bias Compensation, respectively. The green bars show the performance difference between the two conditions.}
\label{fig:ablate_bias}
\end{figure}

In structured pruning of large language models, restoring model performance after the pruning process is a crucial aspect. Our approach uniquely leverages bias compensation as a strategy to recover the performance of pruned models, circumventing the need for expensive and time-consuming retraining procedures. Figure~\ref{fig:ablate_bias} vividly illustrates the performance of the \method method on the WikiText2 dataset, comparing the perplexity scores with and without bias compensation at varying pruning ratios for the LLaMA-7B model. Evident from the figure, bias compensation plays a significant role in mitigating the performance degradation associated with pruning. Furthermore, this compensatory effect becomes more pronounced as the pruning ratio increases, highlighting the growing importance of bias compensation in more aggressively pruned models.

\subsubsection{Robustness to Calibration Samples.} 
\begin{figure}[h]
    \centering
    \includegraphics[width=0.90\linewidth]{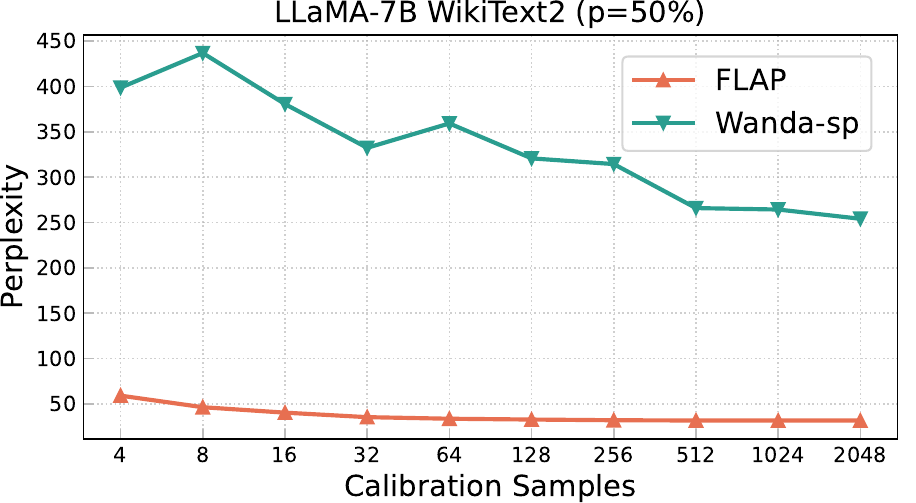}
    \caption{Robustness to Calibration Samples.}
    \vspace{-0.3cm}
    \label{fig:ablate_nsamples}
\end{figure}

Our method utilizes a calibration dataset to estimate the input variance at each layer of the language model. This makes it critical to investigate the impact of the size of this calibration dataset on the pruning performance. Figure~\ref{fig:ablate_nsamples} delineates the effects of varying the number of calibration samples on the pruning outcome. For this analysis, we set a pruning ratio of 50$\%$ for the LLaMa-7B model and observed the resultant perplexity on the WikiText2 dataset. The results clearly show that \method’s performance improves as the size of the calibration dataset increases. In our experiments, we selected a default setting of 1024 calibration samples. Given that only a single forward propagation is required for this calculation, the computational cost associated with this sample size is minimal. Notably, the entire pruning process for the LLaMa-7B model is efficiently completed in a span of 3 to 5 minutes on a single GPU.

\subsection{Inference Speed}\label{sec:exp:speed}

Unlike unstructured pruning, structured pruning offers the dual benefit of reducing both the number of parameters and the inference time, without the need for specialized hardware. This makes structured pruning a more universally applicable approach. In this section, we empirically compare the actual parameter counts and inference speeds of different pruning methods, with the experiments conducted on NVIDIA A100 GPUs. The detailed results are presented in Table~\ref{tab:speed}. Notably, Wanda, employed here as a representative of unstructured pruning, does not effectively reduce either the parameter count or the inference speed. In contrast, our method demonstrates substantial efficiency improvements: at a 20$\%$ pruning ratio, it reduces the number of parameters by 52$\%$, and accelerates the inference speed by 66$\%$. At a 50$\%$ pruning ratio, these improvements are further amplified, with reductions in parameter count by 25$\%$, and an increase in speed by 31$\%$.

\begin{figure}[t]
\vspace{-1mm}
    \centering %
  \includegraphics[width=0.9\linewidth]{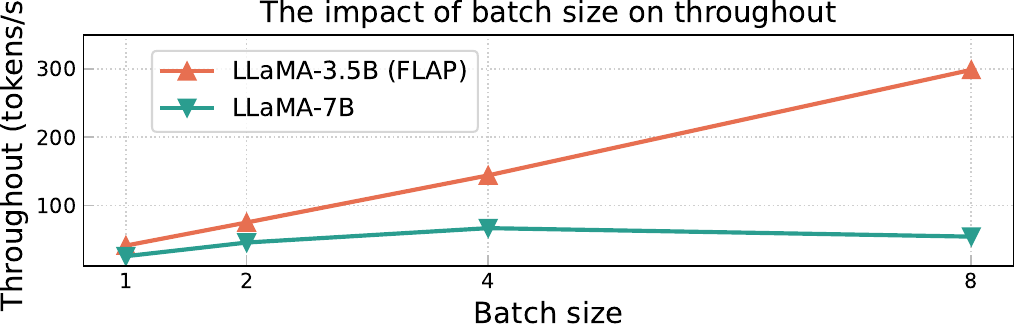}
\vspace{-0.23cm}
\caption{The impact of batch size on throughput. The hardware is the NVIDIA A100-40G.}
\label{fig:batch_size}
\vspace{-3mm}
\end{figure}

Figure~\ref{fig:batch_size} compares the throughput of the LLaMA-7B model with a model pruned by 50$\%$ using our method, across various batch sizes. The comparison clearly shows that the pruned model benefits more at larger batch sizes, as it has not yet hit the throughput bottleneck.

\begin{table}[t]
    \centering
    \resizebox{\linewidth}{!}{
    \begin{tabular}{l|c|ccc}
        \toprule
        Method & Pruning Ratio & Params & Memory & Tokens/s \\
        \midrule
        LLaMA-7B  & 0$\%$ & 6.74B & 12916.5MiB & 25.84 \\   
        \midrule
        Wanda & \multirow{3}{*}{20$\%$} & 6.74B & 12916.5MiB & 25.67 ($\approx$ 0$\%$) \\
        LLM-Pruner &  & 5.42B & 10387.2MiB & 32.57 ($\uparrow$ 26$\%$) \\
        \method (Ours) &  & 5.07B & 9726.2MiB & 33.90 ($\uparrow$ 31$\%$) \\
        \midrule
        Wanda & \multirow{3}{*}{50$\%$} & 6.74B & 12916.5MiB & 25.95 ($\approx$ 0$\%$) \\
        LLM-Pruner &  & 3.35B & 6547.1MiB & 40.95 ($\uparrow$ 58$\%$) \\
        \method (Ours) &  & 3.26B & 6268.2MiB & 42.88 ($\uparrow$ 66$\%$) \\    
        \bottomrule
    \end{tabular}
    }
    \caption{Evaluation results of the inference speed before and after pruning.} 
    \label{tab:speed}
    \vspace{-1mm}
\end{table}

\section{Conclusion}\label{sec:conclusion}

In this work, we propose \methodlong, a retraining-free structured pruning framework explicitly designed for Large Language Models (LLMs). To address the challenges posed by structured pruning, we introduce a novel structured pruning metric, employ adaptive global model compression strategies, and implement robust compensation mechanisms designed to mitigate potential performance losses. Our empirical results affirm that the structured compression model crafted by \method can maintain perplexity and zero-shot performance without any retraining. Especially worth noting is the efficacy of \method in upholding model performance at both low and medium compression rates. Our work demonstrates that bias compensation can largely replace retraining or parameter efficient fine-tuning (PEFT). We hope that our work contributes to a better understanding of structured pruning and performance recovery of LLMs.

\section{Acknowledgements}\label{sec:acknow}

This work was supported by the National Key R$\&$D Program of China (Grant No. 2021ZD0110400), Beijing Municipal Science and Technology Project (Z231100007423004), Zhejiang Lab (No. 2021KH0AB07), and National Natural Science Foundation of China (Grant No. 62206290, 62276260, 62176254, 61976210, 62076235).

\bibliography{aaai24}

\newpage

\onecolumn

\appendix

\section{A \quad Detailed Experimental Settings} \label{app:exp:set}

\subsubsection{Models.} We evaluate \method on the LLaMA model family~\cite{touvron2023llama} and Vicuna-7B model. LLaMA is a set of Transformer-based large language models open-sourced by Meta, mainly including LLaMA-7B/13B/30B/65B. Vicuna is an instruction fine-tuned model based on the LLaMA framework, leveraging user-shared conversations for its training. Given the widespread adoption of these models in the open-source community and their foundational role in numerous applications, their compression performance serves as a significant benchmark. We apply our method to all four LLaMA models to illustrate the fitness of \method for different model scales. While our focus is on LLaMA, the versatility of our approach means it can be extended to other Transformer-based LLMs.

\subsubsection{Evaluation.} Following previous work on LLM pruning~\cite{ma2023llmv3,sun2023simple}, we first evaluate the language modeling capabilities of the pruned model on the WikiText2~\cite{merity2016pointer} validation set. We specifically report on the perplexity metric, which gauges a model's predictive accuracy for the sample set. For a more comprehensive evaluation, we employ the EleutherAI LM Harness, a public evaluation benchmark. With it, we evaluate the model's zero-shot performance across seven pivotal common sense benchmarks: BoolQ~\cite{clark-etal-2019-boolq}, PIQA~\cite{Bisk2020piqa}, HellaSwag~\cite{zellers2019hellaswag}, WinoGrande~\cite{ai2:winogrande}, ARC-easy~\cite{allenai:arc}, ARC-challenge~\cite{allenai:arc} and OpenbookQA~\cite{OpenBookQA2018}. We report both the accuracy of each benchmark and the overall average accuracy.

\subsubsection{Baselines.} We compare our method with two previous pruning methods:

\begin{itemize}
    \item Wanda can be viewed as a continuation of OBD~\cite{lecun1990obd} for LLMs, which takes the weight magnitude multiplied by the $\ell_2$-norm of the corresponding input activation as the importance score, and prunes locally within the weights corresponding to each output feature of the current linear layer. We generalize Wanda tto structured pruning by counting the $\ell_2$-norm of each group of weights within the linear layer as the importance score of the whole group and name it as Wanda-sp.
    \item LLM-Pruner is the first structured pruning method for LLMs. This method requires the computation of global gradient information and LoRA fine-tuning.
\end{itemize}

We also attempted to modify SparseGPT~\cite{frantar2023sparsegpt} to structured pruning, but we fail to obtain reasonable results.

\section{B \quad Implementation Details}

Our experiments are performed on an NVIDIA A100 GPU with 40 GB memory. To prune the LLaMA models, we first load them onto GPUs in 16-bit floating-point format. To facilitate the pruning of larger scale models, standardization of the importance metric and threshold filtering are performed uniformly on the CPU, while the remaining procedures (e.g., pruning and recovering) are executed directly on GPUs. For zero-shot tasks, we utilize the evaluation framework available at \url{https://github.com/EleutherAI/lm-evaluation-harness/}.

\subsection{B.1 \quad Wanda for structured pruning}

In order to demonstrate that the unstructured pruning methods of existing Large Language Models (LLMs) are not well-suited for structured pruning, we modify the Wanda~\cite{sun2023simple} metric to create a structured metric. This adapted metric is designed to align with the characteristics of structured pruning, and it takes the following form:

\begin{equation}
\mathbf{S}^{\ell}_{:,j}=\sum_{i=1}^{row}\left|\mathbf{W}^{\ell}_{i j}\right| \cdot\left\|\mathbf{X}^{\ell}_j\right\|_2
\end{equation}

\noindent{where $|\cdot|$ represents the absolute value operator, $\|\mbf{X}_{j}\|_{2}$ evaluates the $\ell_{2}$-norm of $j$-th features aggregated across $N\times L$ different tokens, and the final score is computed by the sum of the product of these two scalar values.}

\begin{figure*}[t]
\centering
\begin{minipage}[b]{\linewidth}
\begin{algorithm}[H]
    \small
    \caption{\small \methodlong}
    \label{alg:FLAP}
    \setcounter{AlgoLine}{0}
    \LinesNumbered
    
    \KwIn{Original model $\mathcal{F}$, calibration samples $\mathcal{D}_t$, weights of the original model $\mathbf{W}^{\ell}$, total pruning ratio $p$}
    \KwOut{Structured pruning mask $\mathbf{M}^{\ell}$, baseline bias $\mathbf{B}^{\ell}_0$, model $\mathcal{F^{\star}}$ after pruning and bias compensation}
    
    \For{$\ell \gets 0$ \KwTo $len(layers)$} {
 
        \textcolor{gray}{\texttt{\# Calculate the importance score $\mathbf{S}^{\ell}_{:,j}$ of each column of weight matrix}}
        
        $\mathbf{S}^{\ell}_{:,j} \leftarrow \sum_{n=1}^N ||\mathbf{X}^{\ell}_{n,j,:} - \overline{\mathbf{X}}^{\ell}_{:,j,:}||_2 \cdot ||\mathbf{W}^{\ell}_{:,j}||_2 $

        \textcolor{gray}{\texttt{\# Standardized score for each weight to make the current $\mathbf{S}^{\ell}_a$ and $\mathbf{S}^{\ell}_m$ comparable}}
        
        $\widehat{\mathbf{S}}^{\ell}_a = (\mathbf{S}^{\ell}_a-\mathbb{E}[\mathbf{S}^{\ell}_a]) /(\mathbb{E}[[\mathbf{S}^{\ell}_a-\mathbb{E}[\mathbf{S}^{\ell}_a]]^2])^{\frac{1}{2}} $,\quad 
        $\widehat{\mathbf{S}}^{\ell}_m = (\mathbf{S}^{\ell}_m-\mathbb{E}[\mathbf{S}^{\ell}_m]) /(\mathbb{E}[[\mathbf{S}^{\ell}_m-\mathbb{E}[\mathbf{S}^{\ell}_m]]^2])^{\frac{1}{2}}$
    }
        \textcolor{gray}{\texttt{\# Generate pruning mask $\mathbf{M}^{\ell}$ by ranking $\widehat{\mathbf{S}}$}}
        
        $\widehat{\mathbf{S}} \leftarrow {\tt\small Concat} (\mathbf{S}^{\ell}_a, \mathbf{S}^{\ell}_m) $, \quad
        $\mathbf{M}^{\ell} \leftarrow {\tt\small TopKMask}(\widehat{\mathbf{S}}, \ p \cdot {\tt\small Size} (\mathbf{M}))$
        
        \textcolor{gray}{\texttt{\# Use the baseline value $\overline{\mathbf{X}}^{\ell}$ to calculate the bias $\mathbf{B}^{\ell}_0$ of Linear layer}}
    
        $\overline{\mathbf{X}}^{\ell}_{:,j,:} \leftarrow \frac{1}{NL} \sum_{n=1}^{N} \sum_{k=1}^{L} \mathbf{X}^{\ell}_{n,j,k}$,\quad
        $\mathbf{B}^{\ell}_0 \leftarrow ((1 - \mathbf{M}^{\ell}) \odot \mathbf{W}^{\ell}) \overline{\mathbf{X}}^{\ell}$,\quad 
        $ \mathbf{W}^{\ell} \mathbf{X}^{\ell} \leftarrow (\mathbf{M}^{\ell} \odot \mathbf{W}^{\ell}) \mathbf{X}^{\ell} + \mathbf{B}^{\ell}_0$

    \Return $ \mathcal{F^{\star}} \leftarrow \mathcal{F}_p(x|\mathbf{M}, \mathbf{W}, \mathbf{B}_0)$
\end{algorithm}
\end{minipage}
\end{figure*}

\subsection{B.2 \quad Pseudo Code}

The detailed steps of our method are outlined in Algorithm~\ref{alg:FLAP}. The inputs to FLAP encompass the original model $\mathcal{F}$, calibration samples $\mathcal{D}_t$, and overall pruning ratios $p$. The final outputs include the global structured pruning mask $\mathbf{M}^{\ell}$, the baseline bias $\mathbf{B}^{\ell}_0$, and the pruned model $\mathcal{F^{\star}}$.

Our approach decomposes the pruning problem for Large Language Models (LLMs) into layer-wise pruning subproblems. In each subproblem, we utilize the calibration data to statistic the input feature information of the corresponding layer and compute the importance score $\mathbf{S}$ using the fluctuation metric. To facilitate the identification of the optimal global compression model structure, we standardize the importance scores of each module in each layer to a standard distribution with a mean of 0 and a variance of 1. Building on this, we merge the importance scores of all layers and modules, then conduct a unified threshold search and to ultimately obtain the global pruning mask. Based on the pruning mask of each layer, we execute the actual pruning of the model and employ the information from the calibration data to introduce additional bias terms, thereby compensating for the reconstruction error of the corresponding layer.

\noindent{\textbf{Streaming update.}}

When counting the information of the input features, we need to estimate the sample mean and sample variance of these features. To minimize repeated calculations and reduce storage, we adopt Welford's method~\cite{welford1962note} to update the mean and variance in a streaming manner. Specifically, the original sample mean and variance are computed using the following formula:

\begin{equation}
\begin{aligned}
    &\bar{x}=\frac{1}{n} \sum_{i=1}^n x_i\\
    &s^2=\frac{1}{n-1} \sum_{i=1}^n\left(x_i-\bar{x}\right)^2
\end{aligned}
\end{equation}

In the context of streaming processing, Welford's method proves to be particularly advantageous as it enables the immediate update of the variance as new samples are received, eliminating the need to recalculate from scratch. This approach is highly efficient, as it not only streamlines data handling but also significantly reduces computational time.

Assuming we have already observed $n$ samples for which we have calculated their mean and variance, when the $n+1$-th sample $x_{n+1}$ arrives, Welford's method allows us to use the following formulas to seamlessly update the mean $\bar{x}_n$ and variance $s_n^2$:

\begin{equation}
\begin{aligned}
\text{Update mean:} \quad &\bar{x}_{n+1}=\bar{x}_n+\frac{x_{n+1}-\bar{x}_n}{n+1}\\
\text{Update variance:} \quad &s_{n+1}^2=\frac{n-1}{n} s_n^2+\frac{\left(x_{n+1}-\bar{x}_n\right)\left(x_{n+1}-\bar{x}_{n+1}\right)}{n+1}
\end{aligned}
\end{equation}

\noindent{\textbf{Group pruning of attention heads.}}

In the structured pruning of Transformers, the self-attention module cannot directly prune the rows or columns of weights. Instead, it necessitates structured pruning at the granularity of the 'head', which is represented by a set of rows or columns of weights. To better align the importance scores between different modules (Self-attn and MLP), we first compute the fluctuation metrics for each column of the weights, doing so separately for each layer and module. We then standardize these metrics, again separately for each layer and module. After standardization, we aggregate the importance scores of neighboring weight columns that correspond to the same head.

In the uniform search across layers and modules, the number of parametric reductions resulting from pruning a self-attention head differs from that resulting from pruning an MLP neuron. To account for this discrepancy, we employ a normalization factor (e.g., 512 / 3). This factor adjusts the importance scores of different modules to be comparable when the same number of elements is removed.

\section{C \quad Additional Experiments}

\subsection{C.1 \quad Zero-shot performance in larger scale}

Table~\ref{tab:llama-13b} presents the zero-shot performance of various downstream tasks with the proposed method applied to the LLaMA-13B model. Our method outperforms the LLM-Pruner, showcasing superior pruning capabilities.

\begin{table*}[t]
    \vspace{-3mm}
    \centering
    \resizebox{0.9\linewidth}{!}{
    \begin{tabular}{l|c|ccccccc|c}
        \toprule
        Method  & Pruning Ratio & BoolQ & PIQA & HellaSwag & WinoGrande & ARC-e & ARC-c & OBQA & Average \\
        \midrule
        LLaMA-13B  & 0$\%$ & 68.47 & 78.89 & 76.24 & 70.09 & 74.58 & 44.54 & 42.00 & 64.97 \\
        \midrule
        LLM-Pruner & \multirow{3}{*}{20$\%$} & 67.68 & 77.15 & 73.41 & 65.11 & 68.35 & 38.40 & 42.40 & 61.79 \\
        LLM-Pruner (w/ LoRA) &  & \tbf{70.31} & \tbf{77.91} & \underline{75.16} & \underline{67.88} & \underline{71.09} & \underline{42.41} & \underline{43.40} & \underline{64.02} \\
        \method (Ours)  &  & \underline{70.12} & \underline{77.59} & \tbf{76.01} & \tbf{69.24} & \tbf{72.59} & \tbf{42.56} & \tbf{43.53} & \tbf{64.52} \\
        \bottomrule
    \end{tabular}
    }
    \vspace{-2.3mm}
    \caption{Zero-shot performance of the compressed LLaMA-13B. \tbf{Bold} results highlight the best performance. \underline{Underscored} results denote the second-best performance for each pruning ratio.} 
    \label{tab:llama-13b}
    \vspace{-6mm}
\end{table*}

\subsection{C.2 \quad Pruning on Vicuna-7B}

In Table~\ref{tab:vicuna}, we present a direct comparison between \method and LLM-Pruner on the Vicuna-7B model, where FLAP demonstrates enhanced performance.

\begin{table}[ht]
\centering
\vspace{-2mm}
\resizebox{0.5\linewidth}{!}{ 
\small 
\begin{tabular}{@{}l|c|ccc@{}}
\toprule
\multirow{3}{*}{Model} & \multirow{3}{*}{Baseline} & \multicolumn{3}{c}{Method (20$\%$ Pruning ratio)} \\
\cmidrule{3-5}
& & \multirow{2}{*}{LLM-Pruner} & \multirow{2}{*}{\shortstack{LLM-Pruner \\ w/ LoRA}} & \multirow{2}{*}{\method} \\
& & & & \\
\hline
Vicuna-7B & 16.11 & 24.70 & 18.97 & \tbf{17.99} \\
\bottomrule
\end{tabular}
}
\vspace{-1.5mm}
\caption{WikiText2 validation perplexity of pruning methods for Vicuna-7B model.}
\label{tab:vicuna}
\vspace{-3mm}
\end{table}

\subsection{C.3 \quad Different calibration data selection}

The selection of calibration data affects the method's generalization. Table~\ref{tab:diff_data} shows that different calibration data suit different downstream tasks, yet the overall average accuracy differences are not significant. Specifically, choosing different calibration datasets, e.g. C4 and WikiText2, results in a fluctuation of about ±1$\%$ in average accuracy for the zero-shot tasks. This phenomenon is also observed in replication experiments with other methods that rely on calibration data, like SparseGPT~\cite{frantar2023sparsegpt}. 

\begin{table}[ht]
    \centering
    \resizebox{0.8\linewidth}{!}{
    \begin{tabular}{l|ccccccc|c}
        \toprule
        Calibration Data & BoolQ & PIQA & HellaSwag & WinoGrande & ARC-e & ARC-c & OBQA & Average \\
        \midrule
        C4 & 60.21 & \textbf{67.52} & \textbf{52.14} & \textbf{57.54} & \textbf{49.66} & 29.95 & 35.60 & \textbf{50.37} \\
        Wikitext2  & \textbf{61.04} & 63.82 & 43.68 & 54.93 & 48.36 & \textbf{30.15} & \textbf{36.31} & 48.33 \\
        \bottomrule
    \end{tabular}
    }
    \caption{The impact of different calibration data on generalization ability.} 
    \label{tab:diff_data}
    \vspace{-3mm}
\end{table}

\subsection{C.4 \quad Generations From Compressed Model}

Table~\ref{tab:generate_llama_5.1b} and Table~\ref{tab:generate_llama_4.5b} present additional examples of the models pruned by our method. We showcase the generation results of both the dense model and the pruned model. Our experiments demonstrate that the pruned LLaMa models with 5.1B and 4.5B parameters, obtained through our pruning approach, are highly effective in retaining general knowledge.

\begin{table*}[t]
    \centering
    \vspace{-2mm}
    \resizebox{\linewidth}{!}{
    \tiny
    \begin{tabular}{p{1.2cm}|p{9.4cm}}
        \toprule
        Model & Generated Sentences \\
        \midrule
        \multirow{21}{*}{\shortstack{LLaMA-7B \\ (baseline)}} & \underline{AI can create a logo in seconds.} In the same way, AI can create a new product or service in seconds. The only thing that is missing is the creative mind of a human. This is why I am so excited about this new technology. It’s going to change the world and create a new era of prosperity for all of us.bolds. The best way to get started is to learn the basics of AI. You can do this by reading my book, Artificial Intelligence for Beginners. It is a short book and easy to read. I have also created a video course to help you learn the basics of AI. This is an excellent way to learn the basics of AI. The course is divided into 10 lessons and takes about 2 hours to complete. You can watch the videos on your computer or mobile device. I hope you enjoy the course and learn a lot about AI. If you have any questions, please leave a comment below. Previous articleWhat is AI? – AI Explained Next articleHow to Learn AI – 5 Ways to Learn AI What is AI? – AI Explained AI is a term that is used to describe a wide range of technologies. In this article, I will explain what AI is and... How to Learn AI – 5 ... \\
        \cmidrule{2-2}
        & \underline{What is McDonald's?} McDonald's is a fast food restaurant that serves hamburgers, french fries, chicken nuggets, salads, milkshakes, soft drinks, and coffee. What is the company's mission? To provide outstanding quality, service, cleanliness, and value to customers. What is the company's vision? To be the world's best quick service restaurant experience by 2015. Who is the CEO? Jim Skinner. How many restaurants does McDonald's have? 31,000 in 119 countries. What are the company's core values? Quality, Service, Cleanliness, and Value. What are the company's strategic objectives? Growth, Innovation, Customer Focus, People, and Financial Discipline. What are the company's financial objectives? To achieve 10$\%$ revenue growth and 15$\%$ EPS growth by 2015. What are the company's marketing objectives? To increase customer satisfaction and loyalty by providing quality products and services at a value that customers perceive as good. What are the company's product/service objectives? To provide a variety of high-quality food and beverages that meet customers' needs and desires. What are the company's pricing objectives? To maintain price leadership by ... \\
        \midrule
        \multirow{15}{*}{\shortstack{LLaMA-5.1B \\ (pruned model)}} & \underline{AI can create a logo in seconds.} In 2017, the AI company Deepgram was able to create a logo that was indistinguishable from the work of a human designer. In 2018, the same company released a video showing how its logo-creation algorithm was able to replicate the style of the world’s most famous graphic designers, including Saul Aalbersheim, Milton Glaser, Massimo Vigna, Paula Parra, and Paul Rigby. AI is a powerful tool for branding and marketing, but it’s not the only thing that can be used to create a brand. Branding can be done by hand, or it can be done with the help of a branding agency. Branding is an art ... \\
        \cmidrule{2-2}
         & \underline{What is McDonald's?} McDonald's is the world's largest fast food chain. It has restaurants in 119 countries and serves over 58 million customers each day. The company's headquarters are in Oakbrook, Illinois, and it employs over 30,000 people worldwide. What is Mcdonalds? Mcdonalds is a chain of fast food restaurants with a menu of hamburgers, cheeseburgers, quarter pounders with cheese, double cheeseburgers, cheeseburger deluxe, big macs, and other food items. How much does Mcdonalds pay? The average salary for Mcdonalds employees is 20,000 per year. The average salary for Mcdonalds employees in the United States is 20,000.SERTIFIED MCDONALD'S INCORPORATION 1000 WEST 10TH STREET, SAN ANTONIO, TX 78216-3400. What is the phone number at Mcdonalds? The phone number for the corporate headquarters of Mcdonalds is 800-355-4378. This number is used to report food ... \\
        \bottomrule
    \end{tabular}
    }
    \caption{Generated Examples from the Compressed LLaMA-5.1B by FLAP.} \label{tab:generate_llama_5.1b}
\end{table*}

\begin{table*}[t]
    \centering
    \vspace{-4mm}
    \resizebox{\linewidth}{!}{
    \tiny
    \begin{tabular}{p{1.2cm}|p{9.4cm}}
        \toprule
        Model & Generated Sentences \\
        \midrule
        \multirow{21}{*}{\shortstack{LLaMA-7B \\ (baseline)}} & \underline{AI can create a logo in seconds.} In the same way, AI can create a new product or service in seconds. The only thing that is missing is the creative mind of a human. This is why I am so excited about this new technology. It’s going to change the world and create a new era of prosperity for all of us.bolds. The best way to get started is to learn the basics of AI. You can do this by reading my book, Artificial Intelligence for Beginners. It is a short book and easy to read. I have also created a video course to help you learn the basics of AI. This is an excellent way to learn the basics of AI. The course is divided into 10 lessons and takes about 2 hours to complete. You can watch the videos on your computer or mobile device. I hope you enjoy the course and learn a lot about AI. If you have any questions, please leave a comment below. Previous articleWhat is AI? – AI Explained Next articleHow to Learn AI – 5 Ways to Learn AI What is AI? – AI Explained AI is a term that is used to describe a wide range of technologies. In this article, I will explain what AI is and... How to Learn AI – 5 ... \\
        \cmidrule{2-2}
        & \underline{What is McDonald's?} McDonald's is a fast food restaurant that serves hamburgers, french fries, chicken nuggets, salads, milkshakes, soft drinks, and coffee. What is the company's mission? To provide outstanding quality, service, cleanliness, and value to customers. What is the company's vision? To be the world's best quick service restaurant experience by 2015. Who is the CEO? Jim Skinner. How many restaurants does McDonald's have? 31,000 in 119 countries. What are the company's core values? Quality, Service, Cleanliness, and Value. What are the company's strategic objectives? Growth, Innovation, Customer Focus, People, and Financial Discipline. What are the company's financial objectives? To achieve 10$\%$ revenue growth and 15$\%$ EPS growth by 2015. What are the company's marketing objectives? To increase customer satisfaction and loyalty by providing quality products and services at a value that customers perceive as good. What are the company's product/service objectives? To provide a variety of high-quality food and beverages that meet customers' needs and desires. What are the company's pricing objectives? To maintain price leadership by ... \\
        \midrule
        \multirow{23}{*}{\shortstack{LLaMA-4.5B \\ (pruned model)}} & \underline{AI can create a logo in seconds.} In 2016, Google’s DeepDream AI was able to create a realistic image of a person from a blank canvas. In 2017, DeepMind, Google’s subsidiary, created a robot that played Go chess, which is a game of checkers that humans have played for centuries. This year, we’re going to see the emergence of artificial intelligence that will be able to do things that we have never dreamed of. The most important thing is that we will be able to create a virtual reality, and it will be more real than our physical world. In the next few years, we will see virtual reality, augmented reality, and holographic displays. These will be used in a variety of industries, including the automotive industry, where they will be the primary driver of the next generation of cars. We will also see virtual reality in sports, where the technology will be used to improve the performance of athletes. In the near future, virtual reality will be a part of our everyday lives. In 2017, the world’s first robotic surgery was performed in the U.S. by robotic arms controlled by doctors from more than 2,000 miles away. Robotics are the fastest-growing segment of the robotics industry, and robot-assisted surg... \\
        \cmidrule{2-2}
         & \underline{What is McDonald's?} McDonald's is a fast food chain that has over 1, locations worldwide. Its headquarters are in Oak Brook, Illinois, and it is the largest fast food restaurant chain in the world. The company operates in more than countries and employs over 60, people. McDonald's serves more than million customers every day, which makes it the most popular fast food chain in the world. The company's revenue in fiscal was billion, and it is the second-largest fast food restaurant chain in the world after Burger King. The company has a large number of restaurants in the United States, but it also has a presence in many other countries, including Canada, the United Kingdom, and Japan. In , the company opened its first restaurant in Japan, and in , it opened its first restaurant in China. In , the company began to serve food in South Korea, and in , it opened its first restaurant in the Middle East. The company also has a large number of franchise locations, which are owned by local businessmen and women. The company is known for its fast food, which is a staple of many diets around the world. The menu includes items such as hamburgers, fried chicken, and burgers, and it has been expanded to include a variety of menu items. The company also sells soft drinks, including Pepsi, Coca Cola ...\\
        \bottomrule
    \end{tabular}
    }
    \caption{Generated Examples from the Compressed LLaMA-4.5B by FLAP.} \label{tab:generate_llama_4.5b}
\end{table*}

\end{document}